\documentclass{article}

\usepackage[preprint,nonatbib]{neurips_2022}

\usepackage[utf8]{inputenc} 
\usepackage[T1]{fontenc}    
\usepackage{graphicx}       
\usepackage[colorlinks = true,
            linkcolor = blue,
            urlcolor  = blue,
            citecolor = blue,
            anchorcolor = blue]{hyperref}\usepackage{url}            
\usepackage{booktabs}       
\usepackage{amsfonts}       
\usepackage{nicefrac}       
\usepackage{microtype}      
\usepackage{xcolor}         
\usepackage[bibstyle=ieee, citestyle=numeric, sorting=nyt, url=false, eprint=false, isbn=false, maxbibnames=99]{biblatex}
\AtEveryBibitem{\clearfield{month}}
\AtEveryBibitem{\clearfield{day}}

\usepackage[stable]{footmisc} 

\usepackage{etoolbox}

\title{EleutherAI: Going Beyond ``Open Science'' \\ to ``Science in the Open''}

\author{
    Jason Phang\thanks{
      Correspondence: \href{mailto:jasonphang@nyu.edu}{jasonphang@nyu.edu}, \href{mailto:stella@eleuther.ai}{stella@eleuther.ai}
    }
    \\ EleutherAI
    \\ New York University
    \And
    Herbie Bradley
    \\ EleutherAI
    \\ University of Cambridge
    \And
    Leo Gao
    \\ EleutherAI
    \And
    Louis Castricato
    \\ EleutherAI
    \\ Brown University
    \And
    Stella Biderman
    \\ EleutherAI
}

\begin{document}

\maketitle

\begin{abstract}

Over the past two years, EleutherAI has established itself as a radically novel initiative aimed at both promoting open-source research and conducting research in a transparent, openly accessible and collaborative manner.
EleutherAI's approach to research goes beyond transparency: by doing research entirely in public, anyone in the world can observe and contribute at every stage.
Our work has been received positively and has resulted in several high-impact projects in Natural Language Processing and other fields.
In this paper, we describe our experience doing public-facing machine learning research, the benefits we believe this approach brings, and the pitfalls we have encountered. 


\end{abstract}

\section{Introduction}

Over the last two years, there has been rapid proliferation of independent AI research groups and non-traditional challengers to the convention of research being conducted primarily at university, government, or corporate labs.
Despite this proliferation, EleutherAI remains (to our knowledge) the only major research lab in the world to do its work primarily in the public eye.
Below, we describe our experience conducting high-impact research in a open-source setting and the significant benefits that this strategy brings, including improved discussions surrounding the research, faster collaboration, engaging participants from non-traditional backgrounds, and fostering a highly active and motivated community of researchers.
We hope that our experience will motivate other groups to consider shifting their research model towards more openness and transparency.


\section{Background on EleutherAI}

\subsection{A Brief History Lesson\footnotemark}
\footnotetext{A more complete early history of EleutherAI can be found at \url{https://blog.eleuther.ai/year-one/}.}

EleutherAI began in July of 2020 with the initial goal of creating a publicly available replication of GPT-3.
The early members of EleutherAI were largely not from the ML or NLP academic communities, but consisted mostly of software engineers, ML hobbyists, and researchers in fields outside of machine learning, and many had non-traditional backgrounds.
Participation was driven by interest in ensuring that a similarly capable large language model would be available for public use, study, and scrutiny, both to support open research as well as to ensure that the dangers and pitfalls of such models would not be hidden behind closed doors.

The main line of EleutherAI's language modeling work resulted in the creation and public release of the GPT-Neo 1.3B and 2.7B \cite{black2021gpt}, GPT-J-6B \cite{gpt-j}, and GPT-NeoX-20B \cite{black-etal-2022-gpt} models, each of which were the largest publicly available decoder-only English language models at their time of release\footnote{A version of Megatron-LM (11B) was released prior to both GPT-Neo and GPT-J-6B, but severely underperformed even much smaller models and is generally excluded from such discussions.} and the last of which was the largest publicly available English language model of any type. This work also resulted in the creation of other resources now commonly used by many language modeling groups: The Pile \cite{gao2020pile}, an 800GB diverse pretraining corpora for LMs, and the LM-Eval-Harness \cite{eval-harness}, a toolkit for evaluating language models' zero- and few-shot performance on a diverse set of NLP tasks.

Beyond the work on language modeling, the scope of EleutherAI's research has also greatly expanded over time, driven by the growth of the community and the varied interests of its members.
Today, EleutherAI's research efforts cover a range of topics, including model interpretability (as-of-yet unpublished), multimodal models \cite{vqgan-clip}, and large model training infrastructure \cite{oslo,gpt-neox-library}. EleutherAI has also recently spun up CarperAI, a research lab within EleutherAI focused on reinforcement learning from human feedback \cite{ouyang2022training,bai2022training}.


Additionally, EleutherAI is uniquely one of the few open ML research organizations deeply involved in AI safety, with a focus on AI alignment and other AI risks since its inception.
A section of EleutherAI is dedicated to in-depth technical alignment discussion, commentary on recent publications, and reading groups for beginners in the field. 




\subsection{Organization and Collaboration}


EleutherAI is organized entirely as a Discord server that can be freely joined by anyone.
Notably, there is no formal notion of ``membership'' in EleutherAI---participation is entirely voluntary and membership is via self-identification.
The server contains two main types of channels: a set of general discussion channels, ranging from casual conversation to more technical research discussion, and a set of channels for coordinating ongoing and active projects.
Aside from a small number of private channels for administrative and moderation purposes, all channels are open and immediately accessible to all members.
EleutherAI operates on a public-by-default convention.

While the small number of discussion channels can lead to cross-talk and chaotic discussions, we have found that it has also lead to active discussion, cross-pollination of ideas, and greater awareness of ongoing efforts by other members, thereby creating more opportunities for collaboration.
As such, EleutherAI resembles a digital interpretation of a  research lab or working space where anyone can wander the halls and listen to others as they work.

%

\subsection{The Project Development Pipeline}

EleutherAI projects are organized in an entirely grassroots manner, started by members interested in studying a research question or creating a given model or resource. Members then pull in volunteers and acquire computational resources to execute a given project as needed. The catalyst for new projects tends to be newly released work, newly announced findings, or research questions spawned from an existing project.

Initial experiments are often run on an individual basis, but once sufficient momentum and interest in a project has been established, members often recruit volunteer research collaborators to assist with further development.
When the project is sufficiently scoped out, members may request and draw on EleutherAI's shared computational resources, such as GPU allocations or cluster access.
A dedicated channel may also be created for a project to facilitate more technical and in-depth discussion.
Because of the public-by-default nature of the EleutherAI server, the project discussion channel serves a secondary purpose as an informal ``newsletter'', where updates and discussions can be read by other members who drop by the channel out of interest.
They may in turn offer help in troubleshooting, sharing expertise, or providing other resources to assist with the project.
Finally, completed projects that have significant EleutherAI participation are announced to the whole server. 




\section{Public-Facing Research as a Mission}


Since its inception, EleutherAI has adopted a principle of doing public-facing research, where all discussion is held publicly by default, and where there are minimal barriers to newcomers passively spectating or actively contributing to research discussion and research efforts.

EleutherAI has seen remarkable success not only in executing high-impact research projects, but also in fostering an active community with high quality research discussion while being coordinated entirely online with participants from around the world. We believe that the public-facing and open nature of EleutherAI, a quality found in almost no other research lab, has contributed to this success.

While public-facing research is closely connected to EleutherAI's goal of making open-access models, our discussion here is focused on the open nature of collaboration within EleutherAI and how it affects the research process, and not on the open-access end products of projects.


\subsection{Public-Facing Research Enables Conversation}

We have found an array of benefits from being public-facing that we did not originally anticipate. For example, by conducting high-impact research in a public-facing setting, where both high-level planning and low-level implementation details are accessible to all, we have been able to cultivate a tremendously active community with quality research discussion.
This in turn attracts other researchers who may be tackling similar challenges in their own work, who chime in with their own suggestions, experiences, and anecdotes, thereby proliferating the high level of technical discussion.

EleutherAI has also fostered a generally lighthearted tone of discourse, partially owing to the use of the Discord platform, which is generally associated with gaming and entertainment communities.
Hence, researchers feel more comfortable sharing informal details about their work such as implementation specifics, partially substantiated conjectures, and negative results. The latter are particularly valuable and often lacking in more traditional research settings.


Importantly, almost all research discussion and project coordination in EleutherAI occurs on Discord, with minimal usage of other synchronous media such as video calls or voice chat.\footnote{
There are a small number of recurring voice chat sessions, including reading groups, alignment discussion groups, and casual weekly voice chat.}
This means that almost all discussion is publicly accessible and searchable, and can be carried out asynchronously by people on opposite sides of the globe. EleutherAI's history of conversation and research efforts provides newcomers with a full and transparent perspective of how research is conducted, the community norms, and the culture of work, making it easy for newcomers to get caught up on discussion or project progress and then contribute to an ongoing effort with minimal friction.

Lastly, being an entirely online organization, participation in EleutherAI is open to virtually anyone with an Internet connection.
With participation from hobbyists and academic researchers alike from around the globe, EleutherAI serves as a round-the-clock research discussion and collaboration hub.



\subsection{Public-Facing Research Fosters Collaboration}

EleutherAI's public-facing collaboration model has many similarities with open-source software development, which has been a highly effective model on projects including the development of the Linux operating system \cite{foss-development-raymond}, and which relies on public discussion of bugs and feature development.

The core principles of open-source software (OSS) development can also be applied to research---open research collaborations can be a powerful tool for accelerating scientific discovery.
By sharing data, code, and results in public, collaborators can build on each other's work more effectively and quickly identify problems and potential solutions. Just as in OSS development, the public nature of projects provides a built-in mechanism for quality control as the community can provide feedback and suggestions for improving ideas, code, and papers: \textit{given enough eyeballs, all bugs are shallow}. 

Public-facing research also minimizes the barriers to entry for a potential new contributor, ensuring that it is easy for newcomers to access and become familiar with project details and progress, and to start contributing to an ongoing effort. New and aspiring researchers can be easily absorbed into the community as collaborators, since they can learn from and contribute to existing projects before proposing their own. The quick and easy onboarding of both collaborators and ideas also means that the pipeline from discussion to execution is extremely short.

In practice, we have found that this speed in turn encourages more serious research discussion since participants know that discussion is likely to turn into experiments quickly, forming a virtuous cycle. 
By providing a space for researchers and researchers to share model implementations and iterative improvements, new ideas can build off previous ones at an accelerated pace. 
For example, the publicly shared text-to-image VQGAN-CLIP \cite{vqgan-clip} Colab attracted significant community interest and involvement, leading to many variants and tweaks improving on the initial implementation.



\subsection{Public-Facing Research Creates Community}


Beyond the practical benefits, the public-facing nature of EleutherAI has also allowed us to create a vibrant and supportive community of members.
Members feel included and supported in the process of doing research when active discussion is held in public, and even members who are not actively involved in the project feel welcome to ask questions and offer suggestions.
Members can thus freely adjust their level of involvement in EleutherAI efforts while still feeling like a part of the EleutherAI community.
In fact, many members do not participate in any EleutherAI project, but still frequent the channels contributing to discussion, sharing knowledge, and chiming in with suggestions.

EleutherAI's public discussion norms are particularly valuable for aspiring researchers from non-traditional or non-academic backgrounds, who may lack an established learning path or clear guidance on how to get involved in ML research.
Anecdotally, we have observed cases of individuals with minimal computer science backgrounds and no prior research knowledge becoming core contributors or leading research projects within a matter of weeks.
The welcoming of curious and enthusiastic new entrants to the field feeds back into EleutherAI's culture, reinforcing the norm of embracing individuals of disparate levels of expertise and varied backgrounds.




\section{Downsides and Limitations}

Despite the many benefits to the public-facing, volunteer-based operation of EleutherAI discussed above, we have also experienced some challenges that accompany this model.

\paragraph{Volunteer-based Participation}
Participation in EleutherAI efforts is voluntary and without direct financial compensation, and maintaining ongoing commitments from participants has been a  challenge.
Members initially enthused by a project may passively reduce their involvement over time, fail to deliver on assigned work, or do so with significant delays, thereby slowing down a project.
EleutherAI has been fortunate to have had both highly motivated members and high-profile projects which attract significant interest and effort from participants, but in the long run, finding ways to encourage, maintain and reward high-quality contributions is crucial for continued success.

\paragraph{Scooping}
The public-facing nature of research and project discussion provides ample opportunities for ideas and findings to be taken from EleutherAI efforts, and published or publicized without attribution to EleutherAI's members.
This is a significant concern given the rapid pace of research and publication within the ML and NLP fields, as well as the possibility for industry labs or well-funded academic groups with more resources to execute certain ideas more quickly. While we have taken some steps to correct the public record, such as citing our own tweets (e.g., \textcite{biderman2021gee}) and documenting findings on our blog that we do not write up formally, this is not always sufficient. Ultimately, we have decided that this is an acceptable sacrifice to make for the substantial good we have experienced.

\paragraph{Not Everything Should Be Public} 
Another potential concern of public-facing research is the dissemination of information that has net negative effects on the world, such as research that is dual-use.
While we apply serious consideration to the releases of our major research artifacts,\footnote{\url{https://blog.eleuther.ai/why-release-a-large-language-model/}, \textcite{black-etal-2022-gpt} (Appx C)} the very nature of public-facing, open research means that we cannot apply similar scrutiny to all discussion.
Here, we again rely on community norms to ensure that information is shared responsibly, keeping in mind that the research we are doing can have a material impact on society and on the development of AI systems which might contribute to catastrophic or even existential risks.

\section{Conclusion}

Over twenty years ago, forty machine learning researchers resigned from the editorial board of the paid-access \textit{Machine Learning} journal (MLJ) to support the open-access \textit{Journal of Machine Learning Research} (JMLR), citing the need for research to be freely available for the community.\footnote{\url{https://en.wikipedia.org/wiki/Machine_Learning_(journal)}}
This mass resignation in favor of open publishing was a watershed event in the modern history of machine learning research, and demonstrated the community's firm commitment to openness of research.
We quote from their resignation letter:\footnote{\url{http://sigir.org/files/forum/F2001/sigirFall01Letters.html}}

\begin{quote}
    We believe, however, that there is much more to achieve, and that our further growth and further impact will be enormously enhanced if via  [JMLR] we are able to communicate more freely, easily, and universally.

    $\cdots$

    In summary, our resignation from the editorial board of MLJ reflects our belief that journals should principally serve the needs of the intellectual community, in particular by providing the immediate and universal access to journal articles that modern technology supports, and doing so at a cost that excludes no one.
\end{quote}

In an era where cutting-edge research is increasingly driven by proprietary models and datasets developed opaquely by industry labs, EleutherAI is proud to continue the tradition of commitment to open research, in both product and process.

EleutherAI has succeeded beyond the wildest expectations of its early members, and we attribute this success in large part to our commitment to doing research in public, which has allowed us to collaborate with talented individuals from varied backgrounds and maintain a vibrant community of researchers.
While this recipe may not be applicable to all organizations, we nevertheless urge the community to move toward conducting more research and discussion in the open.

\newpage

\printbibliography

\newpage

\appendix

\section{Acknowledgements}
EleutherAI has enjoyed generous support and contributions from several parties, including both financial support and computation resources. We would like to extend our utmost gratitude to CoreWeave, TPU Research Cloud, StabilityAI, The Eye, Lambda Labs, and Nat Friedman for their generous support.

\section{Comparison with Other Open Research Labs}

In Section 1, we claim
\begin{quote}
    Despite this proliferation, EleutherAI remains (to our knowledge) the only major research lab in the world to do its work primarily in the public eye.
\end{quote}

This by no means is meant as a dismissal of the work of groups like BigScience or LAION, but does represent a substantial philosophical difference. BigScience is organized via an invite-only Slack that non-participants cannot gain access to. LAION has a public discord server but makes extensive use of private threads and has a second entire private discord server where a substantial amount of their research occurs before it is ready for public consumption. The ML Collective's mission focuses on community building research training, and functions more like a hub for connecting like-minded researchers than a lab in-and-of itself. 

A further point of differentiation is that none of these groups are as involved in AI safety and especially technical AI alignment as EleutherAI is. While these organizations are to some extent involved in AI safety---for instance LAION has released an NSFW detector and BigScience has licensed their releases under the RAIL licence taking into consideration potential harms of LLMs---there is little to no focus on technical AI alignment and AI existential risk in these groups, and consideration of these factors is not as pervasive as in EleutherAI. In addition to facilitating alignment discussion, EleutherAI has published a series of blog posts about technical alignment, largely produced out of the alignment discussions in the server. EleutherAI has also hosted an alignment reading group with its own curriculum (to be published) forked from the Cambridge AGI safety fundamentals course.

\end{document}